\documentclass{article}

\PassOptionsToPackage{numbers}{natbib}

\usepackage[preprint]{neurips_2022}

\usepackage[utf8]{inputenc} 
\usepackage[T1]{fontenc}    
\usepackage{hyperref}       
\usepackage{url}            
\usepackage{booktabs}       
\usepackage{amsfonts}       
\usepackage{nicefrac}       
\usepackage{microtype}      
\usepackage{xcolor}         
\usepackage{relsize}
\usepackage{amsmath}
\usepackage{mathtools}
\usepackage{array}
\newcolumntype{P}[1]{>{\centering\arraybackslash}p{#1}}
\usepackage{tabu}
\usepackage{upgreek}

\title{Self-Supervised Learning of Phenotypic Representations from Cell Images with Weak Labels}

\author{%
  Jan Oscar Cross-Zamirski\\
  DAMTP \& AstraZeneca\\
  University of Cambridge \\
  Cambridge CB3 0WA \\
  \texttt{jc856@cam.ac.uk}
  \And
  Guy Williams \\
  Discovery Sciences, R\&D \\ 
  AstraZeneca \\
  Cheshire, SK10 4TG \\
  \texttt{guy.williams2}$^{*}$
  \And
  Elizabeth Mouchet \\
  Discovery Sciences, R\&D \\ 
  AstraZeneca \\
  Cheshire, SK10 4TG \\
  \texttt{elizabeth.mouchet}$^{*}$
  \And
  Carola-Bibiane Schönlieb\\
  DAMTP \\
  University of Cambridge \\
  Cambridge CB3 0WA\\
  \texttt{cbs31@cam.ac.uk} 
  \And
  Riku Turkki $^{\dag}$\\
  Discovery Sciences, R\&D \\
  AstraZeneca \\
  Gothenburg, 431 50 Mölndal \\
  \texttt{riku.turkki}$^{*}$ 
  \And
  Yinhai Wang $^{\dag}$\\
  Discovery Sciences, R\&D \\
  AstraZeneca \\
  Cambridge, CB4 0WG \\
  \texttt{yinhai.wang}$^{*}$
  \And
  $^{*}$\texttt{@astrazeneca.com}
   \And
  $^{\dag}$\textnormal{Equal contribution}
  }
\begin{document}

\maketitle
\begin{abstract} 
We propose WS-DINO as a novel framework to use weak label information in learning phenotypic representations from high-content fluorescent images of cells. Our model is based on a knowledge distillation approach with a vision transformer backbone (DINO), and we use this as a benchmark model for our study. Using WS-DINO, we fine-tuned with weak label information available in high-content microscopy screens (treatment and compound) and achieve state-of-the-art performance in not-same-compound mechanism of action prediction on the BBBC021 dataset (98\%), and not-same-compound-and-batch performance (96\%) using the compound as the weak label. Our method bypasses single cell cropping as a pre-processing step, and using self-attention maps we show that the model learns structurally meaningful phenotypic profiles. 
\end{abstract}

\section{Introduction}
Deep learning methods for phenotyping cells from microscopy images have advanced rapidly alongside cellular imaging technology \citep{pratapa}. Profiles captured from images can be used to quantify activity and perform downstream tasks such as mechanism of action (MOA) prediction of cells treated with compounds of interest. These methods hold promise for accelerating drug discovery pipelines \citep{chandrasekaran}.


Typically, convolutional neural networks (CNNs) have been the method of choice, however these models can struggle to generalise to new treatments, and can be limited by label information \citep{lukraus}. Rarely are they interpretable, and this may contribute to how machine learning methods have not reached their potential in clinical use \citep{clinical}. Across medical imaging fields, vision transformers (ViTs) \citep{vit} and self-supervised \cite{ssl} methods have risen as a viable alternative to CNNs with labels \cite{swav} \citep{johan}.


In this study we make the following contributions:
\begin{itemize}
    \item We introduce Weakly Supervised DINO (WS-DINO), a framework to incorporate weak label information into a self-supervised knowledge distillation construction based on the DINO algorithm \citep{dino}.  
    \item We implement WS-DINO, and with the BBBC021 dataset \citep{BBBC021} \citep{BBBC021_labelled} we learn representations with two weak labels - treatment and compound. We achieve state-of-the-art results for not-same-compound (98\%), and not-same-compound-and-batch (96\%) mechanism of action prediction using the compound as the weak label. We show the learnt representations are biologically meaningful and based on cellular structure with self-attention maps.
\end{itemize}



\section{Background}
\label{gen_inst}

\subsection{Image-based profiling}

Imaged-based profiling \citep{chandrasekaran} is the process of extracting features from high-content images (HCI) \citep{hcs} of cells. The features extracted from these kinds of images are used to build profiles which can assess bioactivity and MOA of the cells in response to treatments with chemical compounds and/or genetic perturbations. In drug discovery these phenotypic profiles are used to compare new treatments with known ones, and downstream applications include routine lead compound identification, drug target screening through CRISPR technology and toxicity assessment \cite{marianna}.

Most phenotypic representation learning methods are based on single cell segmentation \citep{perakis} \citep{ando_2017}. When applied to large datasets this can lead to high computational demands in pre-processing images with segmentation algorithms \citep{cellprofiler}. Working with single-cells requires aggregation to the population level and may necessitate further computational requirements for corrections due to cellular heterogeneity \citep{rohban}. In high-content imaging there is metadata which can be considered a \textit{weak} label, examples being treatment compound and concentration. In previous work a classification network was trained predict the weak label as an auxiliary task, with a frozen layer of the network forming the phenotypic representation used in downstream tasks such as MOA prediction \citep{Caicedo2018_weak}.

Unsupervised methods such as self-supervised learning constructions \cite{ssl} using a contrastive loss \cite{contrastive} have been very successful. Self-supervised algorithms such as SwAV \citep{swav} and DINO \citep{dino} give consistently richer embeddings for downstream tasks compared to a pretrained supervised baseline model in medical imaging \citep{truong}. Self-supervised models have performed very well learning feature represenations from single cell microscopy images \citep{lukraus} \citep{perakis}, and recently on the image-level with a multi-scale approach \citep{janssens}.

\subsection{DINO - self-supervised learning with knowledge distillation}

DINO from Caron \textit{et al.} \citep{dino} is a self-supervised learning approach which incorporates aspects of knowledge distillation \citep{knowledge_distil}. Uniquely combining self-supervised learning with a Vision Transformer (ViT) \citep{vit} backbone, DINO can learn features shown to perform well at \textit{k}-NN clustering tasks. 

DINO consists of  student network $g_{\theta_s}$, parameterized by $\theta_s$, trained with set of image crops to match the teacher network $g_{\theta_t}$, parameterized by $\theta_t$, which sees a different set of crops. For each image $x$ a set of views $V$ is generated which contains the two global crops $x^{g, 1}$ and $x^{g, 2}$ as well as eight local crops. The student is passed the set of global and local crops $V$, while the teacher sees only the global crops. The student network is trained to maximise the agreement between the outputs of $g_{\theta_s}$ and $g_{\theta_t}$. To achieve this, the parameters of $\theta_s$ are found by minimizing the cross-entropy loss: 
\begin{center}
\begin{equation}\min\limits_{\theta_s}  \mathlarger{\mathlarger{\sum}}\limits_{x\in \{x^{g, 1}, x^{g, 2} \}}   \; 
\mathlarger{\mathlarger{\sum}}\limits_{\substack{x' \in V \\ x' \neq x}} H(P_t(x),P_s(x')) \end{equation} 
\end{center}
Where $H(a, b) = - a \log b$ and $P_s$ and $P_t$ are the probability distributions of the student and teacher respectively (defined in the Appendix). Teacher weights $\theta_t$ are frozen during each epoch of student training and updated iteratively with an exponential moving average based on the previous weights of the student network with the formula: $\theta_t \leftarrow \lambda\theta_t + (1-\lambda)\theta_s$, where $\lambda$ is the momentum parameter. 

Both the student and the teacher network have a ViT \cite{vit} backbone. The inputs are small patches of fixed size ($N \times N$) in a non-overlapping grid, which are embedded (alongside a positional embedding) with a linear layer. An additional learnable [CLS] token with a projection head output $h$ is added to the embeddings and sent to the transformer encoder. The attention mechanism allows the ViT to synthesise information across the whole image as self-attention layers in the Transformer globally update the attention token embeddings.

\section{Methods}
\label{headings}



\subsection{WS-DINO}

We propose a \textbf{W}eakly \textbf{S}upervised form of self-\textbf{di}stillation with \textbf{no} labels (WS-DINO) as an adaptation to the DINO algorithm. The aim of the approach is to retain the benefits of self-supervised learning with knowledge distillation while also incorporating the weak label information as part of the training. 


We introduce the notation $x_{i, y_i}$ to represent the $i^\text{th}$ field of view of a fluorescent channel in the dataset which has been treated with treatment or compound $y_i$ - the weak label. The superscript contains the crop information: $g$ for global and $l$ for local crops. When generating the sets of different views $V_t$ (seen by teacher) and $V_s$ (seen by student) for training, we enforce the constraint that the global and local crops are sampled from different images with the same weak label. We define the sets $V_t$ and $V_s$ for the randomly sampled ordered pair $(i,j)$ where $i\neq j$ and $y_i = y_j$:
\begin{center}
\begin{equation}
V_t =  \{x_{i, y_i}^{g,1}, x_{i, y_i}^{g,2} \}
\end{equation}
\begin{equation}
V_s =  V_t \cup \begin{rcases} \begin{dcases}
      x_{j, y_j}^{l,k} : & k \in  \{ 1, ... , n \} \\
    \end{dcases}   \end{rcases}     
\end{equation}
\end{center}
Where the superscript details the $k^\text{th}$ local crop of $n$ total crops (default: $n=8$). Sampling the local crops from a different image is in contrast to sampling random crops with different augmentations from the same image, which is the method of DINO. For WS-DINO we minimize the loss:
\begin{center}
\begin{equation}\min\limits_{\theta_s}  \mathlarger{\mathlarger{\sum}}\limits_{x\in V_t}   \; 
\mathlarger{\mathlarger{\sum}}\limits_{\substack{x' \in V_s \\ x' \neq x}} H(P_t(x),P_s(x')) \end{equation}
\end{center}
In this section we have presented the key details of our adaptation to the DINO algorithm from the original paper \citep{dino}. We provide implementation details in section 3.4 and in our GitHub repository. \footnote{\url{https://github.com/crosszamirski/WS-DINO}}



\subsection{Dataset and augmentation}

We evaluated our experiments with the publicly available BBBC021 dataset \citep{BBBC021}, which consists of MCF-7 breast cancer cells exposed to several chemical compounds for 24h. The cells are stained with three fluorescent labels: DNA, F-actin, and $\upbeta$-tubulin. For training and evaluation, we used the subset which has been labelled for MOA evaluation \citep{BBBC021_labelled} which consists of 12 unique MOA, 38 unique compounds and 103 unique treatments (compound/concentration pairs) across 10 experimental batches. There are 1320 3-channel images in the dimethyl sulfoxide (DMSO) treated control group, and 2528 3-channel images with MOA annotations. Several studies have used the annotated dataset to evaluate classical and machine learning strategies \citep{BBBC021_labelled} \citep{perakis} \citep{ando_2017} \citep{Caicedo2018_weak}  \citep{janssens} \citep{singh} \citep{godinez}   \citep{pawlowski} \citep{lafarge} \citep{goldsborough}.




\textbf{Pre-processing}:  To remove systematic variations in pixel intensities across each field of view, we used a CellProfiler \citep{cellprofiler} pipeline to perform an illumination correction.  A smoothing function with a filter size of $320$ pixels was used to create an illumination correction function. Each image was corrected by dividing all pixel intensities by the illumination correction function. This methodology is consistent with best practice established by the JUMP Cell Painting Consortium \citep{jumpcp}. After correction, images were resized using bicubic interpolation to a size of $640 \times 512$ pixels. A maximum pixel value cut-off of $10,000$ was enforced before normalizing each image to have a mean of $0$ and standard deviation of $1$. The scripts to replicate the pre-processing are available in our GitHub repository.

\textbf{Post-processing}:  Next we take the median feature embedding from four $224 \times 224$ crops around the centre of each image (equivalent to a $448 \times 448$ pixel centre crop split into four non-overlapping crops). To aid in correcting for batch effects (which can be significant in HCI screens) and to capture the range of variation of unperturbed cells, we applied typical variation normalization (TVN) \citep{ando_2017} as a post-processing correction. TVN is a technique where a principal component analysis (PCA) transformation without dimensionality reduction (whitening) is learned using the DMSO embeddings (aggregated to one embedding per image from the median of four crops around the centre). The mean of each of the field-level embeddings is taken across a plate, followed by median aggregation to treatment level resulting in 103 embeddings - in line with the studies following Ando \textit{et al.} \citep{ando_2017}.

\subsection{Performance validation and evaluation}

We evaluated the corrected and aggregated feature embeddings with two metrics ubiquitous in representation learning studies using the BBBC021 dataset: not-same-compound (NSC) matching \citep{BBBC021_labelled} and not-same-compound-and-batch (NSCB) matching \citep{ando_2017} with the MOA labels. NSC matching is a 1-Nearest-Neighbour (1-NN) match for each given treatment to the nearest neighbour in representation space which is \textbf{not} of the same compound. Cosine distance was used as the distance measure. NSCB matching is the same method as NSC matching with the additional constraint to exclude treatments from the same compound \textbf{and} batch as the given treatment. A small number of treatments have a MOA label only present in a single batch and these were excluded in NSCB evaluation.

\subsection{Training}

For all the models the following variables are kept constant: network architecture (ViT-S/8 backbone with 3-layer multi-layer perceptron head), size and number of crops: 2 global $224\times224$ random resize crops (range = $0.1-0.2$ of full size image) and 8 local $96\times96$ random resize crops (range = $0.04-0.08$), teacher momentum = 0.99, and batch size = 16. There was no batch normalization, weight decay or gradient clipping. The optimizer was adamW \citep{adamw}. We applied horizontal and vertical flips (with a probability of 0.5 per sample) as augmentations in training. Learning rate and temperature parameters are controlled with adaptive cosine schedules with linear warm-up. After 10 epochs of linear warmup, the learning rate = $4\times10^{-6}$ and the teacher temperature = 0.04. The learning rate was decayed to $3\times10^{-6}$ over 400 epochs, and the teacher temperature kept constant after warmup.  All other parameters unless otherwise stated are default DINO parameters.  A separate model was trained for each of of the three channels. Every model was initialized with weights from DINO trained on RGB ImageNet \cite{dino} \cite{imagenet} (grayscale images were converted to RGB). Embeddings of size 384 were extracted for each channel then concatenated to a size of 1152 followed by L2 normalization. We provide a full PyTorch implementation in our GitHub repository.


\section{Results and Discussion}




\normalsize

 As a baseline we trained the unsupervised DINO network \cite{dino}. First, we evaluated with ImageNet weights only (transfer learning). Next we fine-tuned the model on the BBBC021 annotated dataset without weak labels. We then trained WS-DINO with the same network and parameters. Two weak labels $y$ were evaluated: treatment and compound. Additionally we trained with MOA as the label $y$, although this cannot be considered a weak label. The images in training were sampled using a weighted random sampler enforcing even sampling of images from each weak label group.
 
 \begin{table}[h]
  \caption{A summary of the results of our experiments. Further details of mean NSC and NSCB scores in training with a discussion of best epoch selection is provided in the Appendix.}
  \label{table1}
  \centering
  \begin{tabular}{cccc}
    \toprule
    \textbf{Model}     & \textbf{Weak Label}   & \textbf{NSC} &  \textbf{NSCB}\\
    \midrule
    \textbf{DINO with ImageNet weights only}    & None     &  91\% & 82\% \\
    \midrule
    \textbf{DINO finetuned on BBBC021}  & None   & 92\% & 95\% \\
    \midrule
    
    \textbf{WS-DINO finetuned on BBBC021}    & Treatment &  92\% & 90\% \\
    \cmidrule(l){2-4}
        & \textbf{Compound}  & \textbf{98\%} & \textbf{96\%}\\
    \cmidrule(l){2-4}
        & MOA    & 100\% & 100\% \\
    \bottomrule
  \end{tabular}
\end{table}

\begin{figure}
  \centering
  \includegraphics[width = 14.0cm]{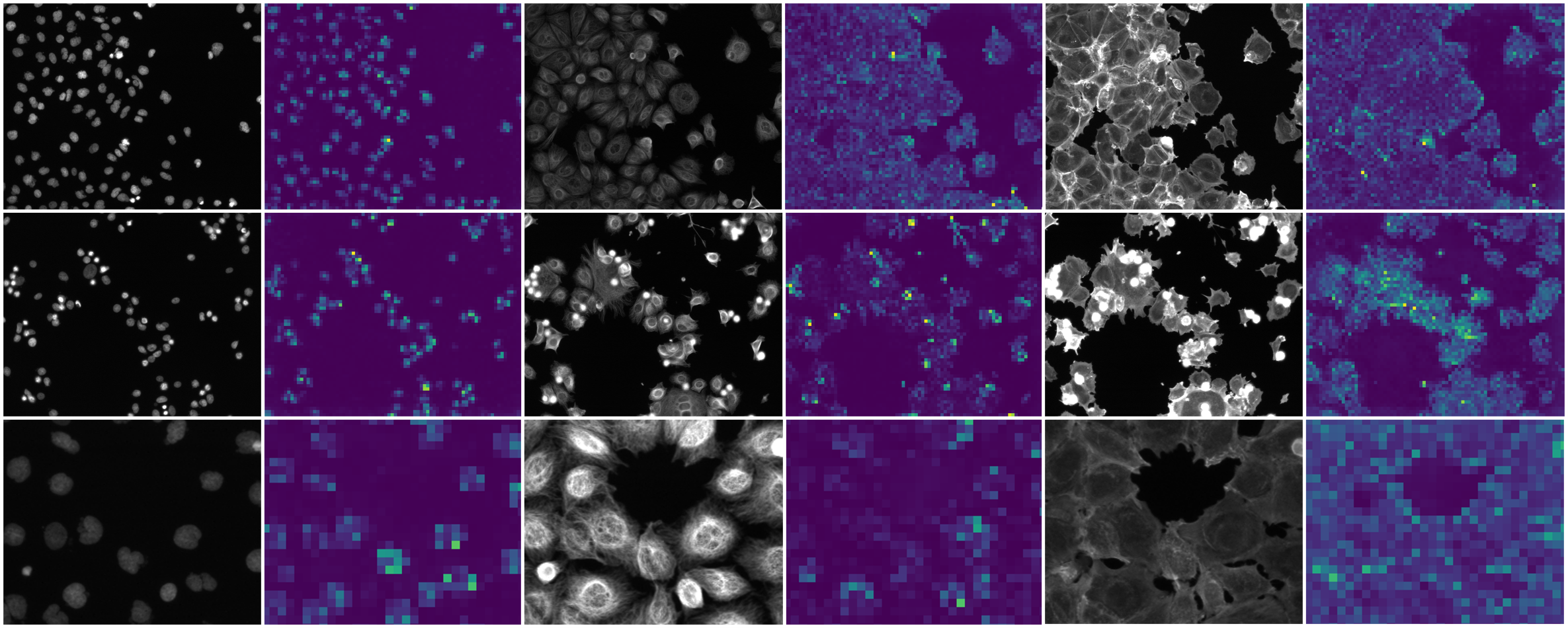}
  \caption{Examples of BBBC021 images with their coupled self-attention maps. Left two columns: DNA channel. Middle two columns: $\upbeta$-tubulin channel. Right two columns: F-actin channel. The first two rows show the images at full size, and the bottom row displays a section of an image zoomed in by a scale factor of four. Produced from WS-DINO weights with compound as the weak label.}
\end{figure}








WS-DINO achieves state-of-the-art results on BBBC021 using the compound as the weak label, outperforming all known previous approaches using this dataset in NSC and NSCB MOA prediction (Table 1 and Appendix). It is notable that using treatment as the weak label does not improve upon the unsupervised approach (DINO). In this dataset all images of the same treatment are from the same batch, however this is not true for the different compounds and MOAs which are expressed with different images across multiple batches. Hence we propose that WS-DINO may provide effective batch correction for datasets with weak label classes with images representing multiple batches. 

 We present self-attention maps (Figure 1) which reveal the segmentation properties of the algorithm. Such visualisations increase confidence in the model by demonstrating the network is learning biologically and structurally meaningful features. 
 
We use MOA as a \textit{psuedo-weak} label as a proof-of-concept for our method. However, we propose future work to evaluate MOA prediction using datasets with partial MOA labels. One advantage of WS-DINO is that it is adaptable to datasets with some known MOA information, a property of many real datasets used in drug discovery. 
 

 WS-DINO synthesises a powerful self-supervised framework with a way to implicitly incorporate the informative weak labels in learning phenotypic representations. Our method can contribute to accelerating drug discovery pipelines by clustering phenotypes in a biologically meaningful hierarchy. The proposed framework is general and the method is not specific to DINO. The sampling of pairs of images with the same weak label is a concept adaptable to other networks. Self-supervised methods are an active field of research and new algorithms have outperformed DINO on ImageNet classification, for example Masked Siamese Networks \cite{msn}. Future work could incorporate these studies which would be adaptable in a very similar way to WS-DINO. Finally, we will expand this research to other datasets with more classes and labels, as well as different downstream tasks. 

\begin{ack}

Jan Oscar Cross-Zamirski is funded by the UKRI-BBSRC DTP (UK Research and Innovation and the Biotechnology and Biological Science Research Council Doctoral Training Partnership) studentship grant 21-3688 and AstraZeneca.

Carola-Bibiane Schönlieb acknowledges support from the Philip Leverhulme Prize, the Royal Society Wolfson Fellowship, the EPSRC grants EP/S026045/1 and EP/T003553/1, EP/N014588/1, EP/T017961/1, the Wellcome Trust 215733/Z/19/Z and 221633/Z/20/Z, the Leverhulme Trust project Unveiling the invisible, the European Union Horizon 2020 research and innovation programme under the Marie Skodowska-Curie grant agreement No. 777826 NoMADS, the Cantab Capital Institute for the Mathematics of Information and the Alan Turing Institute. 

Authors Guy Williams, Elizabeth Mouchet, Riku Turkki and Yinhai Wang are employees of AstraZeneca. AstraZeneca provided the funding for this research and provided support in the form of salaries for the authors, but did not have any additional role in the study design, data collection and analysis, decision to publish, or preparation of the manuscript. We would also like to acknowledge the use of the Scientific Computing Platform (SCP) within AstraZeneca.

\end{ack}

\appendix

\section{Probability distributions of student and teacher}

The probability distributions for the student $P_s$ and teacher $P_t$ are defined by:
$$P_s(x)^{(i)} = \cfrac{\exp{(g_s(x)^{(i)}}/\tau_s)}{\sum_{k=1}^{K}\exp{(g_s(x)^{(k)}}/\tau_s)}$$
for $K$ dimensions and temperature parameters $\tau_s>0$ and $\tau_t>0$ \citep{dino}.

\section{Further training details}

The hyperparameters are fixed across all of the models we train. Many of the parameters are the default values from the original DINO implementation. We optimised the remaining parameters such as learning rate with a small subset of BBBC021 trained without weak labels. In our final models the batch size is limited by GPU memory - a batch size of 16 is small and most likely a larger batch size would improve performance. The models were trained with two GPUs on the AstraZeneca Scientific Computing Platform with a maximum allocation of 32G memory for each GPU.

\begin{table}[h]
  \caption{The results of training with NSC and NSCB scores of the best performing epoch as well as the mean NSC and NSCB scores between 50 and 250 epochs of training each model.}
  \label{table2}
  \small
  \centering
  \begin{tabular}{cccc}
    \toprule
    \textbf{Model}     & \textbf{Weak Label}   & \textbf{Best NSC/NSCB} &  \textbf{Mean NSC/NSCB}\\
    \midrule
    \textbf{DINO finetuned on BBBC021}  & None   & 92\% / 95\% & 90\% / 90\% \\
    \midrule
    
    \textbf{WS-DINO finetuned on BBBC021}    & Treatment &  92\% / 90\% & 89\% / 83\% \\
    \cmidrule(l){2-4}
        & \textbf{Compound}  & \textbf{98\%} / \textbf{96\%} & 96\% / 93\% \\
    \cmidrule(l){2-4}
        & MOA    & 100\% / 100\% & 99\% / 99\% \\
    \bottomrule
  \end{tabular}
\end{table}




We determined the best epoch by following the method of Janssens \textit{et al.} \citep{janssens}. We calculated NSC for each model and selected the epoch with the highest score. This method was chosen to allow a comparison of our model to other studies, however we note there are drawbacks to selecting best epoch with MOA information. In Table 2 the mean values of NSC and NSCB over 200 epochs of training are displayed for each model, and the results show improved performance for NSC and NSCB MOA matching was observed over multiple epochs of training using both compound and MOA as the weak label. The values are lower than the best values as this range contains both under- and over-fit models. We provide epoch data and training logs in our GitHub repository. 

\section{Further results}

\begin{table}[h]
  \caption{A comparison of our method to other selected studies using the BBBC021 annotated dataset.}
  \label{table3}
  \small
  \centering
  \begin{tabular}{cccccc}
    \toprule
    \textbf{Type}     & \textbf{Method}   & \textbf{Reference} &  \textbf{Single Cell} & \textbf{NSC} & \textbf{NSCB}\\
    \midrule
    \textbf{Weakly}  & \textbf{WS-DINO }  & \textbf{This work} & \textbf{No} & \textbf{98\%} & \textbf{96\%} \\
    \textbf{Supervised} & CNN with Mixup & Caicedo \textit{et al.} 2018 \citep{Caicedo2018_weak}& Yes &95\% & 89\% \\
    \midrule
    \textbf{Classical} & CellProfiler &Singh \textit{et al.} 2014 \citep{singh}& Yes & 90\% & 85\% \\
    \textbf{Features}& Factor Analysis &Ljosa \textit{et al.} 2013 \citep{BBBC021_labelled}& Yes & 94\% &77\%\\
    \midrule
    \textbf{Supervised} & Multiscale-CNN & Godinez \textit{et al.} 2017 \citep{godinez}& No & 93\% & N/A\\
    \midrule
    \textbf{Unsupervised}& Contrastive Learning &Perakis \textit{et al.} 2021 \citep{perakis}& Yes & 96\% & 95\% \\
    & UMM Discovery &Janssens \textit{et al.} 2020 \citep{janssens}& No & 97\% & 85\% \\
    & VAE+ &Lafarge \textit{et al.} 2019 \citep{lafarge}& Yes & 93\% & 82\% \\
    & CytoGAN: LSGAN &Goldsborough \textit{et al.} 2017 \citep{goldsborough}& No & 68\% & N/A \\
    \midrule
    \textbf{Transfer} & Deep Metric Network &Ando \textit{et al.} 2017 \citep{ando_2017}&Yes &96\% &95\% \\
    \textbf{Learning} & Inception V3 Pretrained & Pawlowski \textit{et al.} 2016 \citep{pawlowski} & No & 91\% & N/A\\
    \bottomrule
  \end{tabular}
\end{table}

\begin{figure}[h]
  \centering
  \includegraphics[width = 14cm]{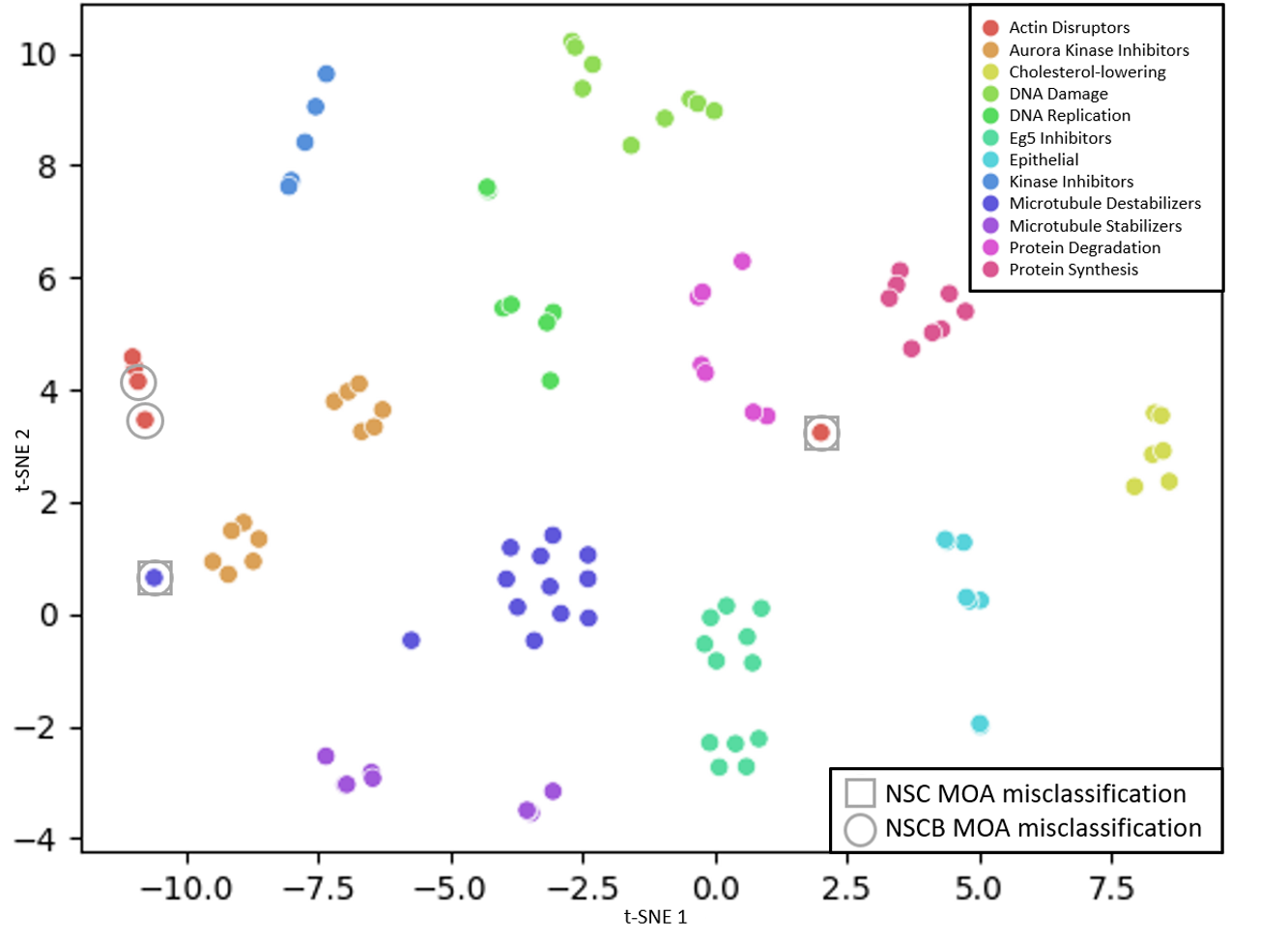}
  \caption{Two-dimensional t-SNE plot of each aggregated treatment feature from 200 epochs of training WS-DINO with compound as the weak label: 98\% NSC and 96\% NSCB MOA classification.}
  \vspace*{4.5in}
\end{figure}


\end{document}